\pdfoutput=1

\documentclass[11pt]{article}

\usepackage[preprint]{acl}

\usepackage{times}
\usepackage{latexsym}
\usepackage{lipsum}
\usepackage{multirow}
\usepackage{subcaption}
\usepackage{tikz}
\usepackage{cleveref}

\usepackage[T1]{fontenc}

\usepackage[utf8]{inputenc}

\usepackage{microtype}

\usepackage{inconsolata}

\usepackage{graphicx}
\usepackage{booktabs}

\usetikzlibrary{shapes.geometric, arrows.meta}
\usepackage{appendix}
\usepackage{tcolorbox}
\usetikzlibrary{shapes,decorations,arrows}

\newcommand{\ellipsesym}{%
  \tikz{\draw[black, fill=gray!45] (0,0) ellipse (0.3cm and 0.15cm);}%
}
\newcommand{\roundrect}{%
  \tikz{\draw[black, fill=gray!15, rounded corners=2pt] (0,0) rectangle (0.6cm,0.3cm);}%
}
\newcommand{\normrect}{%
  \tikz{\draw[black, fill=gray!0] (0,0) rectangle (0.6cm,0.3cm);}%
}

\title{ImmunoFOMO: Are Language Models missing what oncologists see?}

\author{
  \textbf{Aman Sinha\textsuperscript{1,2}} \quad
  \textbf{Bogdan-Valentin Popescu\textsuperscript{2}} \quad
  \textbf{Xavier Coubez\textsuperscript{2}}\\
  \textbf{Marianne Clausel\textsuperscript{1}}\quad
  \textbf{Mathieu Constant\textsuperscript{1}}\\
\\
  \textsuperscript{1}Universit\'e de Lorraine, Nancy, France \quad
  \textsuperscript{2}ICANS, Strasbourg, France
\\
  \small{
    \textbf{Correspondence:} \href{mailto:email@domain}{aman.sinha@univ-lorraine.fr}
  }
}

\begin{document}
\maketitle
\begin{abstract}

Language models (LMs) capabilities have grown with a fast pace over the past decade leading researchers in various disciplines, such as biomedical research, to increasingly explore the utility of LMs in their day-to-day applications. Domain specific language models have already been in use for biomedical natural language processing (NLP) applications. Recently however, the interest has grown towards medical language models and their understanding capabilities. In this paper, we investigate the medical conceptual grounding of various language models against expert clinicians for identification of hallmarks of immunotherapy in breast cancer abstracts. 
Our results show that pre-trained language models have potential to outperform large language models in identifying very specific (low-level) concepts.


\end{abstract}

\section{Introduction}
Language models (LMs) are built for automating linguistic intelligence of humans. In practice, they are evaluated on various natural language processing (NLP) benchmarks including syntactic tasks such as entity recognition or semantic tasks such as inference or document classification. LMs have demonstrated strong performance in domains such as biomedical and clinical text \cite{lewis-etal-2020-pretrained, liu2024largelanguagemodelsclinic}. However, it is crucial for biomedical language models to capture the hierarchical structure of medical concepts (e.g. "pneumonia" being an instance of "respiratory infection" being an instance of "disease").

In the context of oncology research, concepts such as hallmarks of cancer have been proposed to classify the most important principles of cancer development \cite{hanahan2000hallmarks}. 
Immunotherapy in breast cancer is a relatively new addition to the therapeutic arsenal \cite{Loibl2023EarlyBC, Gennari2021ESMOCP}. 
Better understanding the underlying hallmarks of response and resistance to immune checkpoint inhibitors in breast cancer would provide valuable data for developing novel therapeutic strategies and treatment combinations. These hallmarks require nuanced understanding of unstructured clinical or scientific narratives. Despite this, little is known about whether existing language models can reliably recognize and reason such expert-level concepts.

\begin{figure}
    \centering
    \includegraphics[width=\columnwidth, trim = 0.65cm 0cm 0.65cm 0.25cm, clip]{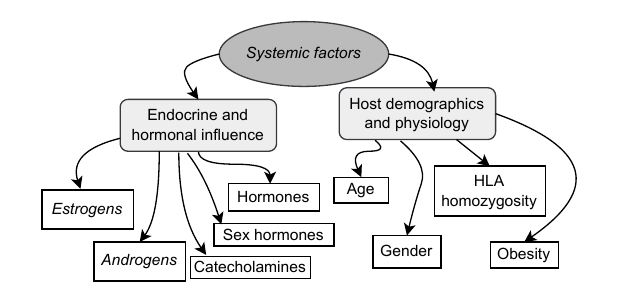}
    \caption{Illustration of conceptual hierarchical structure for a hallmark of immunotherapy (eg. Systemic factors).
    \protect\ellipsesym \ denotes the Tier-I level of hallmark;  \protect\roundrect \ denotes the Tier-II level of hallmark; and \protect\normrect \ denotes the Tier-III level of hallmark. }
    \label{fig:enter-label}
\end{figure}

Previous work in biomedical NLP has largely focused on fact-based question answering, named entity recognition, or relation extraction \cite{BLAKE2010173, yansen_bbac342}. However, far fewer studies have examined the degree to which language models align with expert interpretations in abstract, multi-faceted concept identification tasks — particularly those that demand conceptual inference rather than surface pattern matching \cite{FraileNavarro2025ExpertEO, Workum2025ComparativeEA}. To our knowledge, no existing evaluation benchmark tests model understanding of biomedical conceptual structure as interpreted by clinicians.

In this work, we introduce a novel evaluation study for testing the conceptual grounding of language model against expert-labeled multi-level biomedical concepts in clinical text. We study a range of models—from small language models (SLMs), in particular pre-trained language models (cf. PLMs, e.g., BioBERT \cite{lee2020biobert}, SciBERT \cite{Beltagy2019SciBERT}, BioClinicalBERT \cite{li2022clinical}, etc.) to large language models (cf. LLMs e.g., LLaMA \cite{llama3modelcard}, Qwen \cite{qwen}, Gemma \cite{team2024gemma}) — to identify hallmarks of immunotherapy (HoI) in breast cancer research abstracts. 
In summary, we present a study for evaluating conceptual grounding of language models 
over expert-curated hallmark of immunotherapy concepts. Our study benchmarks a variety of language models and analyzes their agreement with expert annotations. We also identify the limitations of language models' for  capturing complex biomedical concepts.

\section{Related Work}
\paragraph{Language Models in Biomedical domain.}
SLMs'  ability to learn universal language representation led to the creation of domain specific models. One of the pioneer works utilized BERT model architecture by pre-training it on biomedical data to produce BioBERT \cite{lee2020biobert}. This was followed by several other works such as  SciBERT \cite{Beltagy2019SciBERT}, PubmedBERT \cite{gu2021domain}, among others. In parallel, GPT \cite{brown2020language} was introduced. More recently, generative models, including variants of GPT \cite{10.1093/bib/bbac409} and instruction-tuned LLMs  (eg. \cite{zhang2023alpacareinstructiontuned}), have also been adapted for clinical applications. In this study, we utilize two pools of models, one consisting of five PLMs  and the other consisting of five LLMs (See $\S$\ref{sec:methodology} for details). 

\paragraph{Hallmark based text mining.} Hallmarks of cancer \cite{hanahan2000hallmarks} have been  utilized for text mining to identify the trend of cancer research focus of the disease by placing into a fixed set of alterations in cell physiology \cite{hanahan2000hallmarks, baker2017cancer, baker2017initializing}. 
Previous works utilized neural network and language model \cite{baker2017cancer, baker2017initializing} for hallmark classification task and more recently, the rapid advancements in LLMs technologies have inspired new approaches to biomedicine. 

\paragraph{Biomedical conceptual grounding.}
Biomedical language models have shown good performance on tasks related to various medical NLP benchmarks  \cite{yan2024largelanguagemodelbenchmarks}.  Most research focused more explicitly on encoding domain-specific semantic representation by training for sentence or document level tasks which is not directly indicative of preservation of conceptual structures.
We argue that it is crucial, from a medical language understanding perspective, for language models to be able to capture the hierarchical structure of concepts \cite{fivez2021conceptual, khatir2024concept}. We therefore evaluate language models for their capability to capture the hierarchal structure of biomedical concepts. More concretely, we access language models for identification of hallmarks of immunotherapy related concepts that vary from high-level (Tier-I) to low-level (Tier-III) .


\section{Methodology}
\label{sec:methodology}
\paragraph{Dataset.} We extracted free access abstracts from ESMO congresses, namely, ESMO(ES), ESMO-Immunooncology(IO), and ESMO Breast(BR) between 2020-2024 leading to a total of $ \sim$10K abstracts. We then filtered the abstracts in two steps: first, we filtered them to identify breast cancer related abstracts via term (\texttt{breast}) (Filter1). We then applied an additional filter (Filter2) to obtain immunotherapy-centric abstracts via the following four keywords: \texttt{immunotherapy} or \texttt{immune checkpoint} or \texttt{pembrolizumab} or \texttt{keynote} . These two steps resulted in a dataset comprising 239 abstracts. Finally, we also manually verified (MANVER) and discarded any abstract that was not related to breast cancer and/or immunotherapy. We provide the detailed statistics in \cref{tab:esmo-immuno-dataset}. 

\begin{table}
    \centering
    \resizebox{\columnwidth}{!}{
    \begin{tabular}{r rrrr}
        \toprule
         & \textbf{ESMO-IO} &\textbf{ ESMO-ES} & \textbf{ESMO-BR} & \textbf{TOTAL} \\
         \midrule
       All  &810&8232& 967& 10009\\
       +Filter1 & 62 & 1646& 967& 2675\\
       +Filter2 & 27 & 167 & 45 & 239\\
       \midrule
       +MANVER & 26 & 122 & 40 & 188\\
         \bottomrule
    \end{tabular}}
    \caption{ESMO HOI Dataset Description.}
    \label{tab:esmo-immuno-dataset}
\end{table}

\paragraph{Expert Annotations.} Annotations were done by a clinical expert in oncology (notably breast cancer), who provided for each abstract the most appropriate two tier-I hallmark labels. These labels then served as the gold reference for the alignment study of language models and oncologist.

\paragraph{Models.} We used two pools of models: a pool of 5 small language models (SLMs),  and a pool of large language models (LLMs). For SLMs, we utilized BioBERT \cite{lee2020biobert}, ClinicalBERT, ClinicalBigBIRD \cite{li2022clinical}, SciBERT \cite{Beltagy2019SciBERT}, and PubmedBERT \cite{gu2021domain}. For open sourced LLMs, we used Llama-3-8B-Instruct \cite{llama3modelcard}, Gemma-2-9B \cite{team2024gemma}, Med-Qwen2-7B \cite{qwen}, DeepSeek-R1-Distill-Llama-8B \cite{deepseekai2025}, and Biomistral-7B \cite{labrak2024biomistral}. 


\paragraph{Hallmarks.} Each annotated hallmark tier represented a different granularity with the first tier containing 9 hallmarks, the second 27 sub-categories and the last tier 177 relevant keywords. The hallmarks were adapted from literature and clinical experience\cite{morad2021hallmarks, karasarides2022hallmarks, cogdill2017hallmarks}. The list for each tier of hallmarks is provided in the \cref{sec:HoI}.

\paragraph{Task.}  
We performed hallmark identification via unsupervised text classification \cite{Schopf_2022} with all the LMs. For SLMs, we performed similarity based classification \cite{song2014dataless, veeranna2016using}. 
For LLMs, we performed zero-shot hallmark identification task via prompting to the LLM (cf. \cref{fig:prompt-en-llm}). 
We performed the analysis for each of three tiers of hallmarks and evaluated them against Tier-I hallmark labels. More details for reproducibility can be found in \cref{sec:repro}.

\begin{figure}
    \centering
    \includegraphics[width=\columnwidth]{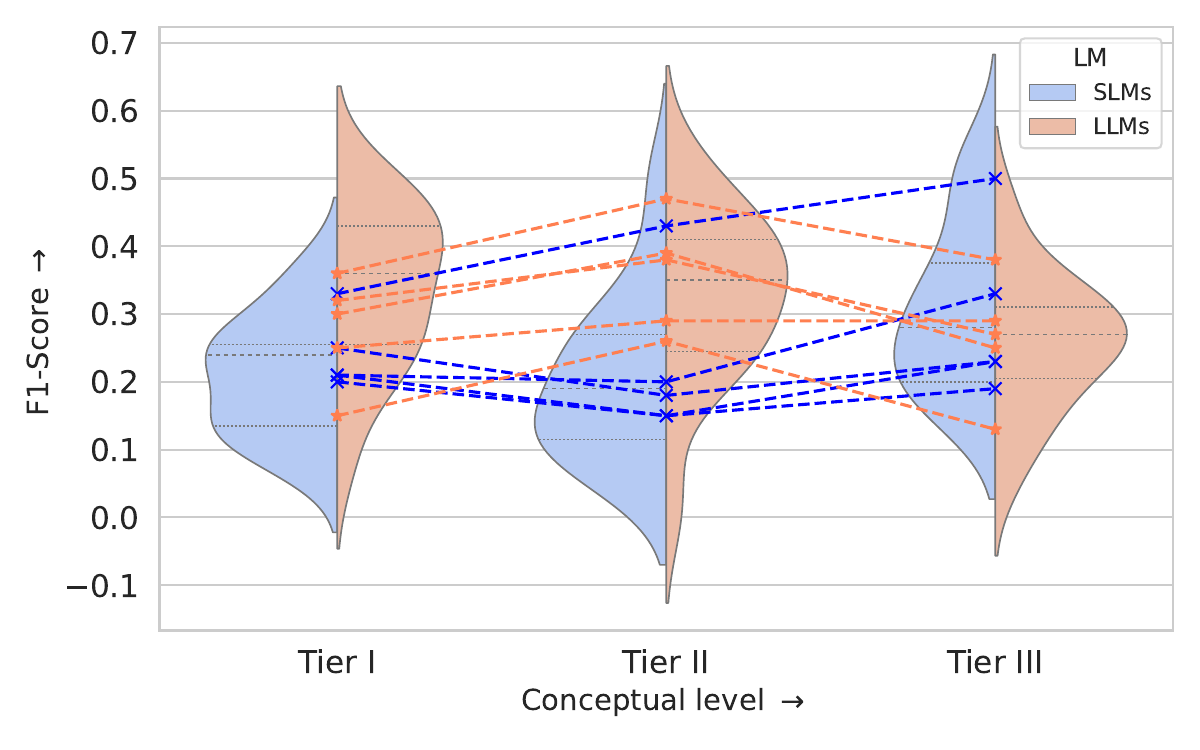}
    \caption{Comparison between SLMs and LLMs against clinical expert for three tiers of concepts of hallmark of immunotherapy. Dashed lines represents each PLMs' (\textcolor{blue}{- -}) and LLMs' (\textcolor{orange}{- -}) overall avg. F1-score on the dataset.}
    \label{fig:against-experts}
\end{figure}

\begin{table}
    \centering
    \resizebox{\columnwidth}{!}{
    \begin{tabular}{c cc c cc c cc}
    \toprule
    & \multicolumn{2}{c}{\textbf{Tier-I}} && \multicolumn{2}{c}{\textbf{Tier-II}} & &\multicolumn{2}{c}{\textbf{Tier-III}} \\
         \midrule
         \multicolumn{9}{c}{Inter-annotator Agreement} \\
         \midrule
         & $C_\kappa$ &  A\% &&$C_\kappa$ &  A\% &&$C_\kappa$ &  A\% \\
         \midrule
        
         SLMs& 0.11&$31.5$& &0.062&$32.1$ &&0.096 &$31.5$\\
         LLMs&0.129& $30.1$ &&0.169&$33.1$ &&0.095& $21.2$\\
         \midrule
         \multicolumn{9}{c}{Intra-annotator Agreement} \\
         \midrule
         & $K_\alpha$ &  C\% &&$K_\alpha$ &  C\% &&$K_\alpha$ &  C\% \\
         \midrule
        SLMs& 0.053&$17.5$& &-0.087&$0$ && 0.006&$0$\\
         LLMs&0.100& $9.6$ &&0.087&$4.9$ &&0.088& $0$\\
         \bottomrule
    \end{tabular}}
    \caption{Alignment of LMs for HoI identification.}
    \label{tab:alignonco}
\end{table}

\paragraph{Evaluation}
We evaluated the models against expert annotations for Tier-I labels\footnote{The curated hallmark list also contained low-level labels in tier-II and III list (See \cref{tab:hallmarks}). So, we converted the lower-level labels (Tier-II or III) to Tier-I based on the aggregation criteria provided in the hallmark list.} 
of HoI using F1-score (weighted avg).
We also evaluated the models for inter-annotator agreement using Cohen's kappa ($C_\kappa$) and percentage agreement (A\%) against the expert and for intra-annotator agreement using Krippendorff's alpha ($K_\alpha$)\footnote{We use Krippendorff's alpha instead of Fleiss kappa to calculate consensus in order to accommodate hallucinations in LLMs which we flag as no predictions.} and percentage consensus (C\%).


\section{Results}


\paragraph{Performance for identification of hallmarks of immunotherapy concepts.} We present comparisons between pretrained language models and large language models against the expert annotations in \cref{fig:against-experts} (See detailed scores in \cref{tab:detailed-scores},\ref{tab:lm-versus-onco}) for identifying the HoI. Among SLMs, PubmedBERT consistently obtained the highest F1-score (weighted avg.) for all the three tiers. In case of LLMs, Gemma-2-9B outperformed other models the pool of LLMs. Overall, for both Tier-I and II, large language models were better than pretrained language models. However, in the case Tier-III, SLMs stood competitive against LLMs. 

\paragraph{LMs' alignment with expert.} We calculated agreement for each language model with the expert by accuracy metric.  Table \ref{tab:alignonco} shows comparison of overall agreement between the pool of pre-trained language models and large language models against the clinical expert respectively. We observed a consistently low agreement between pre-trained language models and clinical experts ($C_\kappa$<0.12). In the case of LLMs, we see a similar low agreement. However, a +3\% increase in overall agreement can be seen toward Tier-II which is followed by a steep 12\% decrease ($C_\kappa$ between 0.9 - 0.17). 

\paragraph{LM's mutual alignment.} We report the overall intra- model pool agreement and consensus percentage for SLMs and LLMs in \cref{tab:alignonco}. Additionally, in \cref{fig:agreement} we present the agreement confusion matrix for every pair of models. 
We observe that there is no consensus in SLMs for lower-tier hallmark labels, LLMs similarly struggle with the Tier-III labels. Both pools of models shows very weak alignment against each other as $K_\alpha$ < 0.1 for all the settings.  

\paragraph{Hallucination in LMs.} We performed strict evaluation for detecting hallucinations by flagging them as any content the model predicted that does not exist in the provided reference list of hallmarks.  We observed no hallucination for the SLMs contrarily to the LLMs. For Tier-I and II, LLMs suffered from roughly 8\%. In Tier-3 however, the percentage increased to 57\%.  


\section{Discussion}

\paragraph{Language models for HoI identification.} The practice of using LLMs via chat interface is prevalent in several domains, and notably in the biomedical setting. This inspired our experiment setting to use language models in an unsupervised manner making it analogous to practical setup. Lower performance was therefore expected for SLMs as they are conventionally trained in a supervised/semi-supervised setting and then tested on downstream tasks. This is reflected in \cref{fig:against-experts} where the overall mean performance is between 15\% to 30\% F1-scores. This is further demonstrated during evaluation of the intra-annotator agreement (See \cref{tab:alignonco}) where the increase in number of classes (from Tier-I to Tier-III) leads to a drop of the SLM model pool consensus to zero for Tiers II and III. However, even with relatively little pre-training data, models such as PubmedBERT and BioBERT performed competitively against LLMs especially for low-level hallmarks confirming their underestimated value compared to the current widespread shift towards large language model usage.

For LLMs, some variation in performance was expected due to the choice of domain specific models such as Med-Qwen or BioMistral, instruction tuned Llama-3B-it versus the rest of the models. Surprisingly, neither instruction tuned nature nor domain specificity was found to be helpful as shown in \cref{tab:lm-versus-onco} where general domain LLMs including Gemma-2-9B performed better than the domain specific and instruction tuned LLMs. Additionally, LLMs did not demonstrate overall success on concept identification task due to their struggle over Tier-III label identification. However, they managed to have better model pool consensus for Tier-II labels compared to SLMs indicating more consistency compared to smaller models.


\paragraph{SLMs are better at low-level concepts, whereas LLMs are better at high-level concepts.} Fig. \ref{fig:against-experts} highlights two consistent trends across the pool of language models. First, the performance of SLMs tends to improve for finer concepts of hallmarks from tier-I (high-level) towards tier-III (low-level) but drops for Tier II. The trend is opposite in the case of LLMs as they perform better for Tier II than Tier I, and then struggle with Tier III concepts. This observation can be attributed to the significant difference between the LMs, with LLMs having a broader coverage of information sources leading them to capture high-level or general concepts better. Then, smaller pre-training data compared to LLMs allows for stronger concept retention and allows models to identify low-level concepts better. 


\paragraph{Hallucinations in LLMs lead to lower precision.} Besides showing the great capability for understanding high-level concepts, large language models tend to produce unwanted or at times incorrect information due to having access to tremendous amount of information sources which can be pre-dominately generic. The higher percentage of hallucinations in the case of Tier-III (low-level) implies that providing an explicit list of potential classes doesn't restrict the LLM to hallmark labels of interest. It was observed that when LLMs are asked to identify low-level HoI for cancer abstracts, they hallucinate producing Tier-II or Tier-I related hallmarks or more generic outputs such as 'immunotherapy' or 'breast cancer'. 

\section{Conclusion}
Conceptual grounding is a crucial criteria for evaluating language understanding capability in language models. In this work, we investigated small and large language models in the context of classifying hallmarks of immunotherapy and evaluated them against a human clinical expert for conceptual grounding. Our study shows that SLMs, in particular domain specific ones, are better at low-level concept identification due to their focused access to domain relevant data and powerful language representation capability. On the other hand, large language models are better at high-level concepts and more consistent compared to SLMs due to having access to general domain prominent data in abundance.  

\section*{Limitations}

\paragraph{Probing intermediate level concepts.}
One of the limitation of this study is the lack of expert annotation at tier-II and tier-III.In the absence of such annotation, we investigated the level of agreement between models as a proxy for the alignment of language models reasoning with human expert.  Such annotations would help better understand the logic behind the abstract tier-I annotation by the human expert. However, the large number of labels for tier-3 (low-level) hallmarks poses difficulties for the LMs to correctly annotate the tier-I hallmarks starting from tier-III.

\paragraph{Filtering of abstracts.}
Adding more filters as opposed to just four ("immunotherapy" ; "immune checkpoint" ; "pembrolizumab" ; "keynote") could increase the  size of dataset and further guiding to robust analyses of potential research strategies for immunotherapy in breast cancer.

\paragraph{Number of tier-III keywords}  Another limitation is the fact that the tier-III list can be further improved with keywords not previously presented. Adding tier-III categories could potentially diminish hallucinations from LLMs as more specific data could be identified and interpreted correctly. We should also note that several keywords can also be reclassified as tier-III for multiple tier-I hallmarks.


\paragraph{Expert annotation} The lack of agreement (see \cref{tab:alignonco}) between models could be reflected in the disagreements between human annotators \cite{Fu2024ExtractingSD}. Therefore, having more annotators could help in explaining the model's disagreement.


\bibliography{custom}

\appendix
\input

\begin{table*}[ht]
    \centering
    \centering
    \resizebox{\textwidth}{!}{
    \begin{tabular}{ c r cccc  r cccc  r cccc}
    \toprule
      && \multicolumn{4}{c}{IO} && \multicolumn{4}{c}{ES} && \multicolumn{4}{c}{BR} \\
      \cmidrule{3-6}
      \cmidrule{8-11}
      \cmidrule{13-16}
        &&  Tier-I & Tier-II & Tier-III & Overall &&  Tier-I & Tier-II & Tier-III & Overall
        &&  Tier-I & Tier-II & Tier-III    & Overall\\
        
         \midrule
       \parbox[t]{2mm}{\multirow{5}{*}{\rotatebox[origin=c]{90}{SLMs}}}  
       &ClinicalBigBird & 0.14&0.11&0.32&0.19 & &0.27&0.19&0.17& 0.21 & &0.11&0.07&0.19 & 0.12\\
        &PubmedBERT& 0.37&\underline {0.37}&\textbf{0.54} & \textbf{0.42} &&0.34&\textbf{0.50}&\textbf{0.49}& \textbf{0.44} &&0.24&0.27&\textbf{0.49} & \underline{0.33}\\
       & BioBERT &  0.11&0.12&0.28 & 0.17&& 0.24&0.25&\underline {0.37}& 0.28 &&0.18&0.12&0.18 & 0.16\\
        &ClinicalBERT & 0.08&0.11&0.32 & 0.17 && 0.24&0.19&0.21 & 0.21 &&0.13&0.07&0.24 & 0.14\\
        &SciBERT & 0.24&0.27&\underline {0.38}&0.29 &&0.29&0.14&0.18 &0.20&&0.16&0.27&0.27 & 0.23\\

        \midrule

         \parbox[t]{2mm}{\multirow{6}{*}{\rotatebox[origin=c]{90}{LLMs}}}  
        &Llama-3-8B-It  & \underline {0.46}&0.33&0.24 & 0.34&&0.36&0.37&0.29& 0.34&&\underline {0.38}& 0.41& 0.23 & 0.34\\
        &Gemma-2-9B & 0.42&0.35&0.33& \underline{0.36}&&\textbf{0.47}&\underline {0.48}&\underline {0.37} & \textbf{0.44}&&\textbf{0.41}&\textbf{0.52}& \underline {0.46} & \textbf{0.46}\\
        &Med-Qwen-7B & 0.31&0.24&0.33& 0.29&&0.28&0.25&0.29 & 0.27&&0.09& 0.02 & 0.27 & 0.12\\
        &DeepSeek-R1 & \textbf{0.50} &\textbf{0.41}&0.18 & 0.36 &&\underline {0.44}&0.37&0.25 & \underline{0.35}&&0.30& \underline {0.42} & 0.27 & 0.33\\
        & BioMistral-7B & 0.23 &0.24&0.06 & 0.17&&0.21&0.28&0.15 & 0.21&&0.22& 0.20& 0.13 & 0.18\\
         \bottomrule
    \end{tabular}}
    \caption{F1-score (weighted avg.) comparison against clinical expert across various congresses.\textbf{Bold} indicates the best performance and \underline{Underline} denotes second best performance.}
    \label{tab:detailed-scores}
    
\end{table*}
\begin{table}[ht]
    \centering
    \resizebox{\columnwidth}{!}{
    \begin{tabular}{ c r cccc}
    \toprule
     
        &&  \textbf{Tier-I} & \textbf{Tier-II} & \textbf{Tier-III} & \textbf{Overall} \\
         \midrule
       \parbox[t]{2mm}{\multirow{5}{*}{\rotatebox[origin=c]{90}{SLMs}}}  
       &ClinicalBigBird & 0.21&0.15& 0.19& 0.18\\
        &PubmedBERT& \underline{0.33}& \underline{0.43}& \textbf{0.50} & \textbf{0.42}\\
        & BioBERT &  0.21& 0.20& 0.33& 0.24\\
        &ClinicalBERT & 0.20& 0.15& 0.23& 0.19\\
        &SciBERT & 0.25& 0.18& 0.23& 0.22\\
        \midrule
         \parbox[t]{2mm}{\multirow{6}{*}{\rotatebox[origin=c]{90}{LLMs}}}  
        &Llama-3-8B-It  &0.32 & 0.38 & 0.27& 0.32\\
        &Gemma-2-9B & \textbf{0.36}&\textbf{ 0.47}&\underline{0.38}& \textbf{0.40}\\
        &Med-Qwen-7B &0.25 & 0.29& 0.29& 0.27\\
        &DeepSeek-R1 &0.30& 0.39& 0.25 & 0.31\\
        & BioMistral-7B &0.15& 0.26& 0.13 & 0.18\\
         \bottomrule
    \end{tabular}}
    \caption{Overall performance of language models against expert across different tiers. \textbf{Bold} indicates the best performance and \underline{Underline} denotes second best performance.}
    \label{tab:lm-versus-onco}
\end{table}

\section{Additional Results}

\subsection{Performance across different tier levels.} 
Table \ref{tab:lm-versus-onco} presents a summary of overall average results for each SLM/LLM model 
 collectively including ESMO-io, es, and br subsets.  \textbf{Overall} column shows that for SLMs, PubmedBERT performs best in the SLM pool of models, obtaining a 42\% F1-Score (weighted avg.). For LLMs, Gemma-2-9B outperforms the other models from the pool by obtaining a 40\% F1-score.

 Table \ref{tab:detailed-scores} presents detailed scores for each SLM/LLM models separately on ESMO subsets datasets (IO, ES and BR) across all the tier levels. 

\subsection{Agreement Confusion matrix.} 
\begin{figure*}[htbp]
    \centering

    \begin{subfigure}[t]{0.32\textwidth}
        \centering 
        \includegraphics[width=\linewidth, trim = 0 0cm 3.8cm 0, clip]{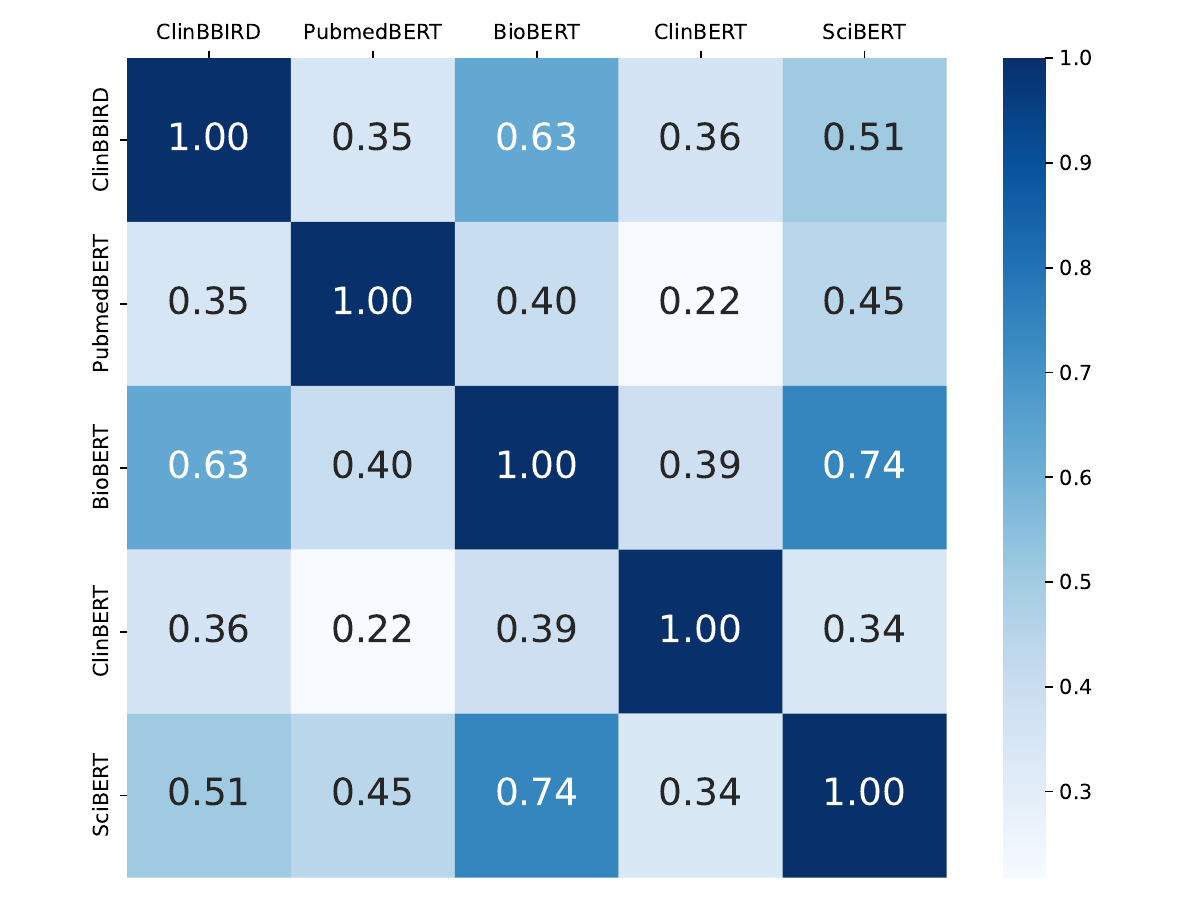}
        \caption{PLM - Tier I}
    \end{subfigure}
    \begin{subfigure}[t]{0.32\textwidth}
        \centering
        \includegraphics[width=0.91\linewidth, trim = 1.5cm 0cm 3.8cm 0, clip]{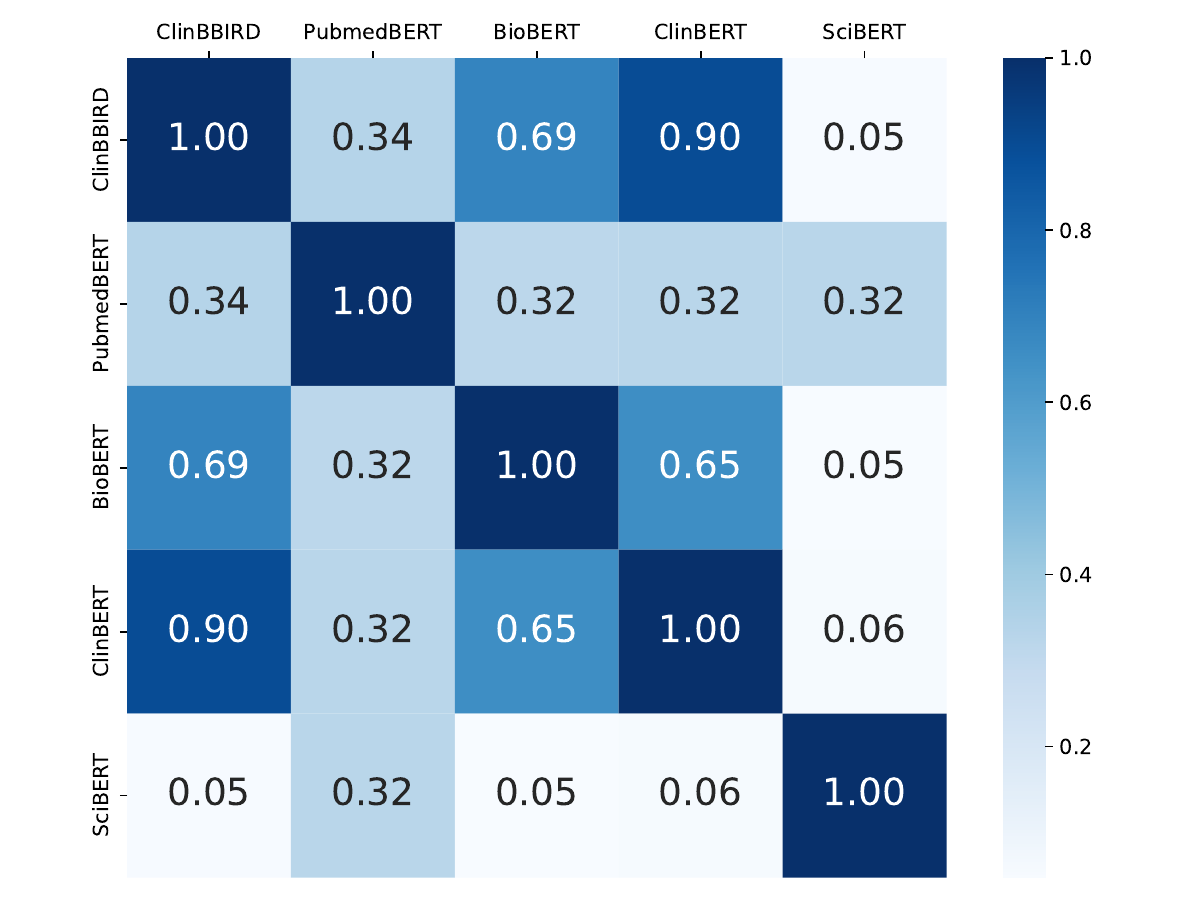}
        \caption{PLM - Tier II}
    \end{subfigure}%
    \begin{subfigure}[t]{0.32\textwidth}
        \centering
        \includegraphics[width=1.15\linewidth, trim = 1.5cm 0cm 0cm 0, clip]{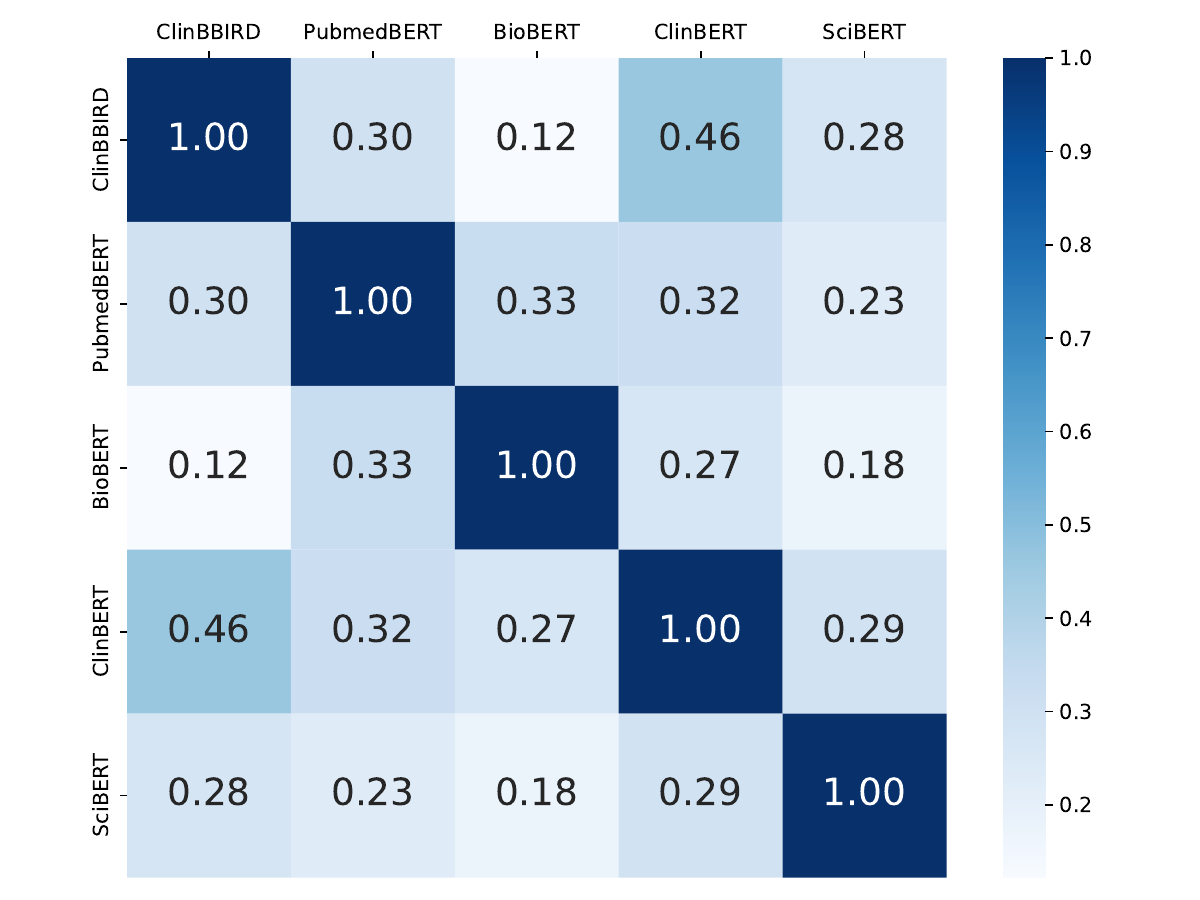}
        \caption{PLM - Tier III}
    \end{subfigure}

    \vspace{0.15cm} 

    \begin{subfigure}[t]{0.32\textwidth}
        \centering
        \includegraphics[width=\linewidth, trim = 0 0cm 3.8cm 0, clip]{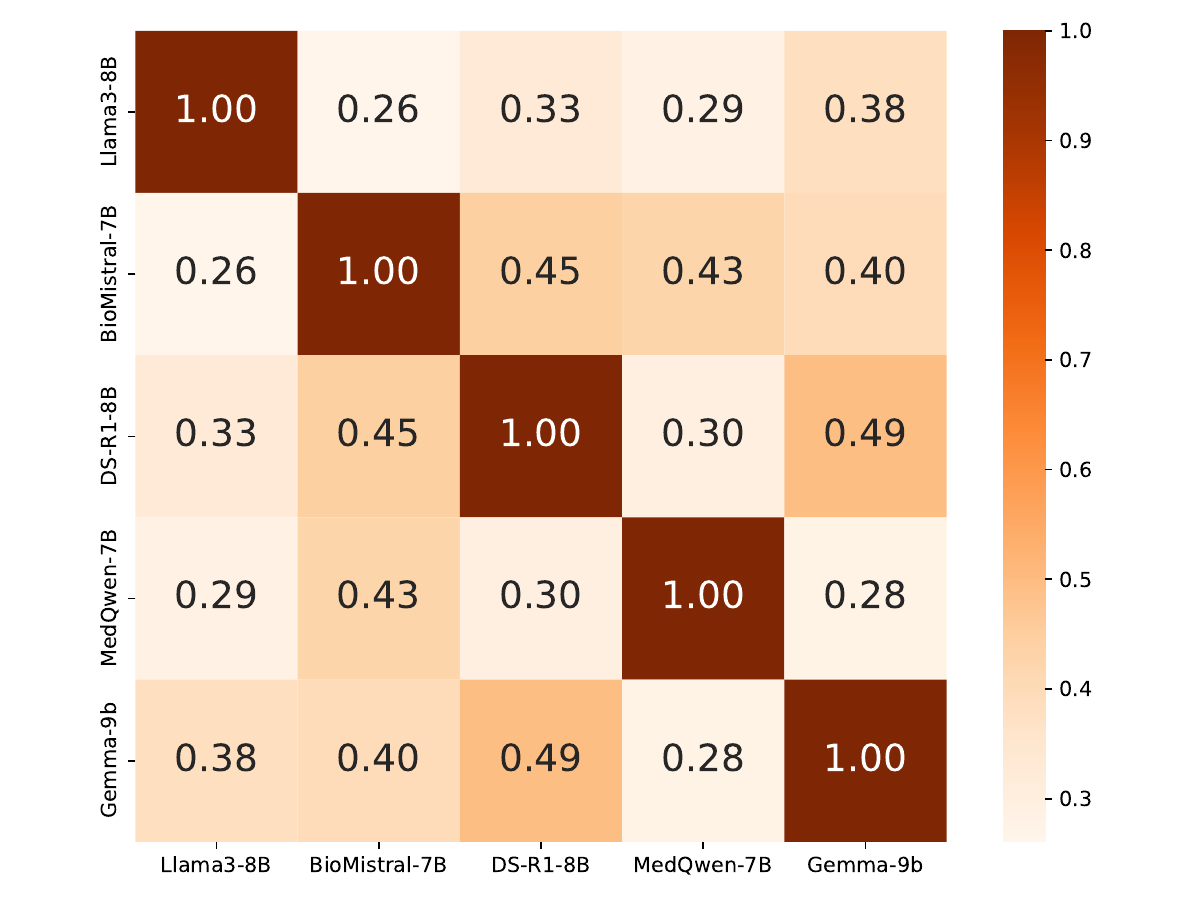}
        \caption{LLM - Tier I}
    \end{subfigure}%
    \begin{subfigure}[t]{0.32\textwidth}
        \centering
        \includegraphics[width=0.91\linewidth, trim = 1.5cm 0cm 3.8cm 0, clip]{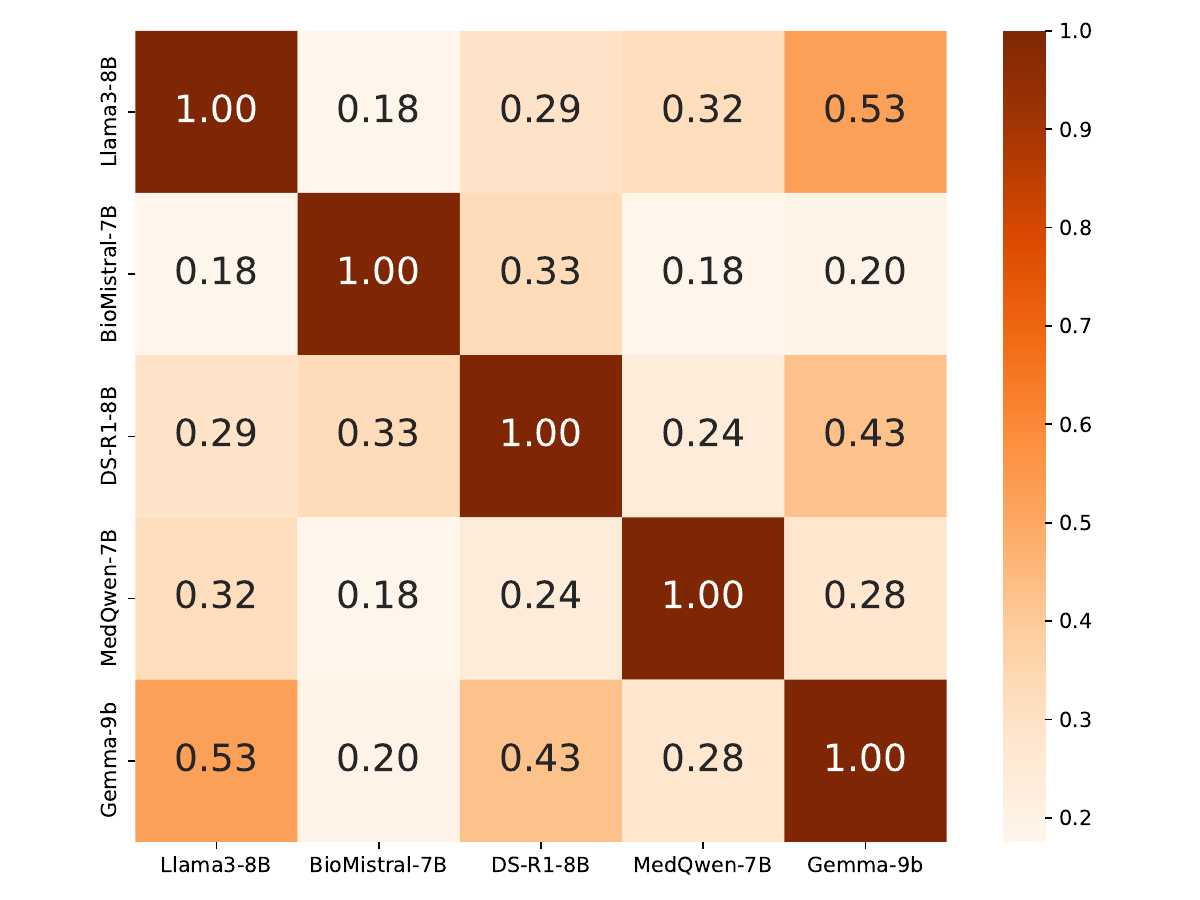}
        \caption{LLM - Tier II}
    \end{subfigure}%
    \begin{subfigure}[t]{0.32\textwidth}
        \centering
        \includegraphics[width=1.15\linewidth, trim = 1.5cm 0cm 0cm 0, clip]{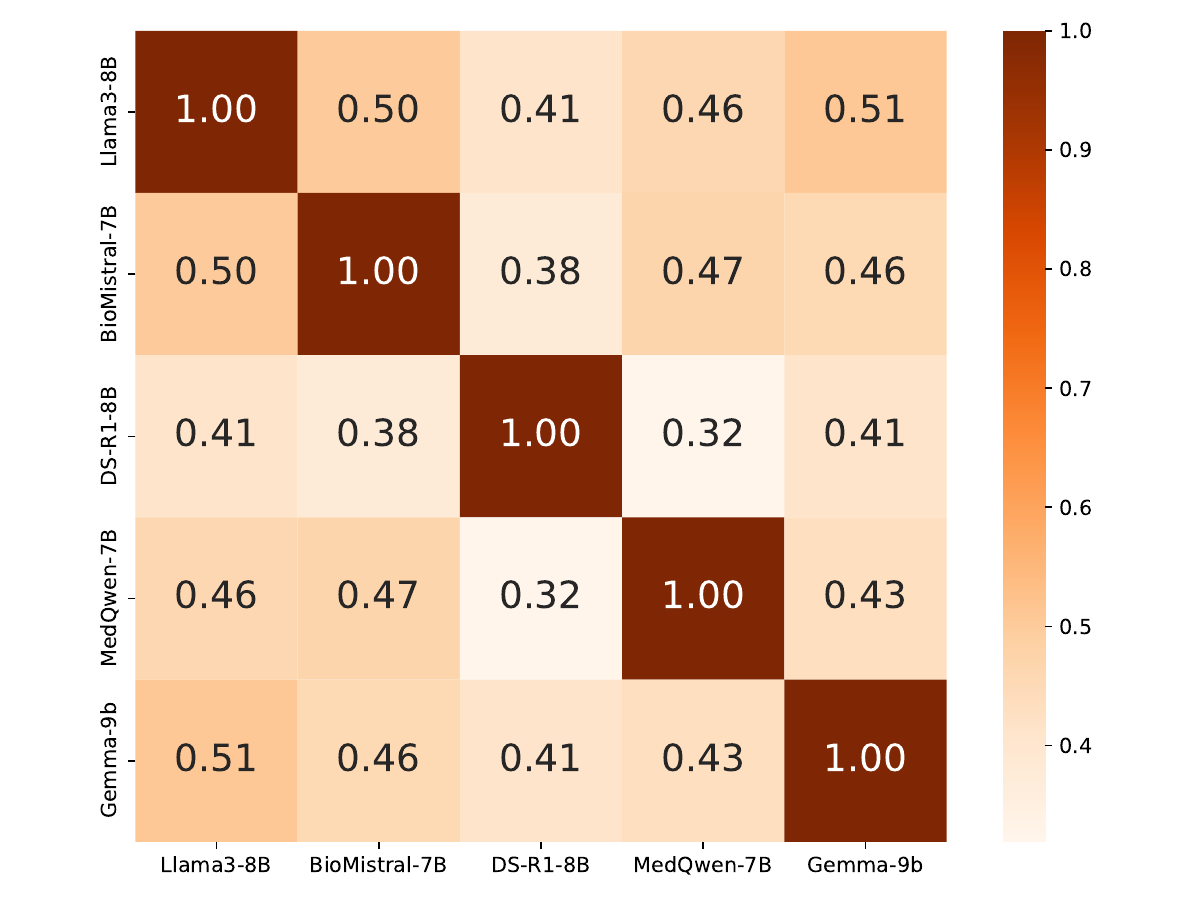}
        \caption{LLM - Tier III}
    \end{subfigure}

    \caption{Intra-alignment (A\%) between LMs over three tiers of HOI concepts (left to right).}
    \label{fig:agreement}
\end{figure*}

Figure \ref{fig:agreement} shows a detailed view of the agreement among the models. In the first row, we show the confusion matrices for the SLM pool with ordering of Tier-I to Tier-III from left to right. In the second row, the same results are shown for the pool of LLMs.

\section{Reproducibility} \label{sec:repro} For pretrained language models, we embedded the given clinical abstract and the given tier list of hallmark labels using the embedding model. We then use utilize the similarity metrics from \texttt{sentence-transformer} library to compute the pair wise score for each hallmark label with the clinical abstract. We select the hallmark label which obtains the highest score from the list of labels.

For large language models, we obtain the predictions using zero shot prompting. For a provided tier list of HoI labels, we prompt the LLM with the prompt template shown in \cref{fig:prompt-en-llm}. Further, we preprocess the answers obtained from the LLM models via GPT-3.5 turbo to obtain the prediction from the raw output generated by the model.
\section{Hallmark of Immunotherapy}
\label{sec:HoI}

We consider a curated list of hallmark of Immunotherapy. 

\paragraph{Tier-I : } The first list contains 9 primary hallmarks -- "Tumor genome and epigenome", "Tumour microenvironment", "Systemic immunity", "Systemic factors", "Microbiome", "Oncogenic signaling", "Tumor metabolism", "Environmental", and "Immune checkpoint inhibitor toxicity".
\paragraph{Tier-II : } The second tier list contains 27 sub-categories. These sub-categories come from the first tier list. 
\paragraph{Tier-III : } The third tier list comprises relevant 177 keywords then fall into one of the 27 categories from Tier-II list.

\begin{figure*}
    \centering
\begin{tikzpicture}
\node[anchor=north west] (box) at (0, 0) {
    \begin{tcolorbox}[colframe=black!75, colback=gray!5, width=\textwidth]
    \small
\textcolor{blue}{Your task is to classify the given abstract based on the following 9 categories :} \\
$[$
    "Tumor genome and epigenome",\\
    "Tumour microenvironment",\\
    "Systemic immunity",\\
    "Systemic factors",\\
    "Microbiome",\\
    "Oncogenic signaling", \\
    "Tumor metabolism",\\
    "Environmental",\\
    "Immune checkpoint inhibitor toxicity" \\
$]$ \\
\textcolor{blue}{ you should return the top three suited categories (in top to bottom order) accurately as a python list. Abstract: \\}

\textcolor{orange}{Durvalumab could be effective in combination with anti-HER2 agents in HER2-low breast cancer (ID 181)	\\
The clinical challenge for treating HER2 (human epidermal growth factor receptor 2)-low breast cancer is the paucity of actionable drug targets. However, the discovery of immune checkpoint inhibitors has made immunotherapy an emerging new treatment modality for breast cancer. Moreover, several chemotherapeutic agents are known to induce immunogenic cell death by activating the immune system. Therefore, we hypothesized that modulating the tumour microenvironment using trastuzumab and or trastuzumab deruxtecan (T-Dxd) in breast organoids co-cultured with T-cells might enhance the response to immunotherapy.	We established a panel of HER2-low breast cancer patient-derived organoids (PDOs), recapitulating the derived tumour. These PDOs were cocultured with immune cells (T- cells and Natural killer cells (NK cells)) and treated with T-Dxd and or trastuzumab in combination with durvalumab. Levels of cytotoxic markers were assessed using flow cytometry and cytokine assays.			Our findings revealed synergistic effects in HER2-low BC patient-derived organoids when treated with T-Dxd and or trastuzumab in combination with durvalumab. We also observed antibody-dependent cellular cytotoxicity (ADCC) response with trastuzumab in combination with durvalumab. These results highlight the need to develop a combination treatment of PD-1/PD-L1 inhibitors with targeted therapies, and other immunotherapies to maximize clinical efficacy.		Altogether, despite preliminary, these findings support the rationale for combining anti-HER2 therapies with immunotherapy in HER2-low BC patients.	
}\\

    \end{tcolorbox}
};
\end{tikzpicture}
\caption{Zero-shot prompting template used for LLM based HOI identification.}
    \label{fig:prompt-en-llm}
\end{figure*}

\begin{table*}[ht]
    \centering
    \resizebox{\textwidth}{!}{
    \begin{tabular}{l p{15cm}}
    \toprule
        \textbf{Tier-I } & \textbf{Tier II}\\
        \midrule
        "Tumor genome and epigenome" & "Genetic alterations and immune visibility",      "Epigenetic dysregulation",      "Antigen presentation and immune evasion"\\
    
        "Tumour microenvironment" & "Immune cell infiltrates",     "Microbiome and extracellular factors",     "Checkpoint molecules and inhibitory receptors",     "Cytokines and soluble immune modulators",      "Immune landscape and spatial architecture"\\
    
        "Systemic immunity" & "Immune cell function and states",     "Immune regulation and dysregulation",      "Soluble mediators and modifiers"\\
    
         "Systemic factors" & "Endocrine and hormonal influence",      "Host demographics and physiology"\\
    
         "Microbiome" & "Microbial sites",     "Microbial imbalances and signatures",     "Microbial species or families"\\
    
         "Oncogenic signaling" & "Key pathways",     "Genetic drivers and regulators" \\
         
         "Tumor metabolism" & "Energy metabolism",     "Metabolic reprogramming"\\
         
         "Environmental" & "Exposures",     "Psychosocial factors",     "Lifestyle",     "Microbial factors"\\
         
         "Immune checkpoint inhibitor toxicity" & "Clinical manifestations",     "Mechanisms",    "Management"\\
        \midrule
        \textbf{Tier-I} & \textbf{Tier-III} \\
        \midrule
        "Tumor genome and epigenome" & 'CTLA-4 polymorphism', 'Human leukocyte antigen (HLA)', 'HLA heterozygosity', 'Beta-2 microglobulin', 'Tumour mutational burden (TMB)', 'Mutational load', 'Mutations/megabase', 'Mutations', 'Neoantigens', 'Genetic alterations', 'DNA mismatch repair', 'dMMR', 'Microsatellite instability (MSI)', 'Histone methyltransferases', 'Dysregulated DUX4 expression', 'Epigenetic alterations', 'DNA methyltransferase inhibitors', 'Histone deacetylase inhibitors (HDAC)', 'Epigenetic chromatin modifications', 'Histone methyltransferases', 'Epigenetic alterations', 'DNA methyltransferase inhibitors', 'Histone deacetylase inhibitors (HDAC)', 'Epigenetic chromatin modifications', 'PALB2', 'RAD51', 'Tumor mutational burden'\\
    "Tumor microenvironment" & 'CD8 T cells', 'CD4 T cells', 'Tumor-infiltrating lymphocyte (TILs)', 'Teff', 'Effector T cells', 'NK cells', 'B cells', 'Tumor-infiltrating B cell', 'Tertiary lymphoid structures (TLS)', 'Treg', 'Tumor-infiltrating Tregs', 'Myeloid-Derived Suppressor Cells (MDSCs)', 'Tumour-associated macrophages (TAMs)', 'M1/M2 macrophages', 'Neutrophils', 'Cancer associated fibroblasts (CAF)', 'Fibroblast activation protein (FAP)', 'Tissue-specific stromal', 'Tumor endothelium', 'Combined Positive Score (CPS)', 'Intratumoral microbes', 'Intratumoral microbiome signature', 'Tumor microbiota', 'Tumor-associated microbes', 'Extracellular vesicles (EVs)', 'Exosomal PD-L1', 'Extracellular matrix', 'PD-1', 'PD-L1 (B7-H1)', 'PD-L2 (B7-DC)', 'CD273', 'CTLA-4', 'CD125', 'LAG-3 (CD223)', 'TIM-3', 'TIGIT', 'VISTA / VSIG', 'B7-H3', 'BTLA (CD272)', 'Siglec-15', 'VEGF', 'IL-2', 'IL-6', 'IL-10', 'IL-12', 'IL-17', 'IL-35', 'TGF-$\beta$', 'Indoleamine-2,3-dioxygenase', 'CSF1R', 'Immune excluded', 'Immune depleted', 'Angiogenesis', 'Cellular components promoting and inhibiting immunity'\\
    "Systemic immunity" & 'Peripheral immune cells', 'CD8+ T-cell immune memory', 'Effector T cells', 'T-cell activation', 'T-cell exhaustion', 'Antigenic stimulation', 'Antigenic tolerance', 'Immunosuppressive phenotypes', 'Dysfunctional IFN-$\gamma$ signaling', 'Costimulation', 'Priming of immune response', 'Peripheral antigen presenting cells (APCs)', 'Host polymorphisms', 'Chronic inflammation', 'Cytokine markers', 'Soluble factors', 'Stress hormones', 'Glucocorticoids'\\
    "Systemic factors" & 'Sex hormones', 'Estrogens', 'Androgens', 'Hormones', 'Catecholamines', 'Age', 'Gender', 'Obesity', 'HLA homozygosity'\\
    "Microbiome" & 'Gut microbiota', 'Oral microbiota', 'Circulating microbiota', 'Dysbiosis', 'Microbiome signature', 'Bacterial ‘signatures', 'Fecal microbiota transplantation', 'Bacteroidetes', 'B. fragilis', 'Bacteroides', 'Burkholderiaceae'\\
    "Oncogenic signaling"  & 'Interferon (IFN)', 'IFN$\gamma$ signaling', 'JAK-STAT', 'PI3K', 'MAPK', 'Wnt/$\beta$-catenin pathway', 'Hedgehog pathway', 'BRAF', 'FGFR3', 'PTEN', 'PPAR-$\gamma$', 'KRAS', 'Myc', 'BRCA', 'HER2', 'Hormone-receptor signaling', 'Estrogen-receptor (ER) signaling', 'Progesteron-receptor (PR) signaling'\\
    "Tumor metabolism" & 'Glucose', 'Lactate', 'ATP', 'NAD+', 'Aerobic glycolysis', 'Oxidative phosphorylation', 'Deregulated tumor immunometabolism', 'Hypoxia', 'Oxidative stress', 'Dysregulated mitochondrial biogenesis'\\
    "Environmental"  & 'Food and water contamination', 'Chemical and industrial exposure', 'Household exposures', 'Air pollution', 'UV radiation', 'Pets', 'Racial injustice', 'Depression/anxiety', 'Psychological and mental stress', 'Socioeconomic inequalities', 'Sexual discrimination', 'Smoking', 'Drug use', 'Medication', 'Surgeries', 'Physical activity', 'Alcohol consumption', 'Diet and dietary supplements', 'Viral antigens', 'Pathogenic microbes', 'Household microbial contamination', 'Environmental non pathogenic microbes'\\
    "Immune checkpoint inhibitor toxicity"& 'Immune toxicity', 'irAE (immune-related adverse effects)', 'Autoimmune disease', 'Aberrant T cell activity', 'Macrophage infiltrate', 'Underlying autoimmune disease', 'Use of systemic corticotherapy', 'Corticotherapy' \\
         \bottomrule
    \end{tabular}}
    \caption{Curated list of hallmarks of immunotherapy. The top half presents a mapping between Tier-I to Tier-II concepts of hallmarks; and the bottom half presents a mapping between Tier-I to Tier-III concepts of hallmarks.}
    \label{tab:hallmarks}
\end{table*}

\end{document}